%% file: main.tex
\documentclass[conference]{IEEEtran}
\IEEEoverridecommandlockouts
\usepackage{graphicx}
\usepackage[backend = bibtex,
			style = numeric,
			sorting = none,
			]{biblatex}
			
{\footnotesize
\bibliography{references.bib}}
\usepackage{xcolor}
\usepackage{comment}
\usepackage{varwidth}
\usepackage{subfig} 
\usepackage{enumitem}
\usepackage{caption}
\usepackage{multirow}

\begin{document}
\author{
    \IEEEauthorblockN{Abhinav Goel, Caleb Tung, Sara Aghajanzadeh, Isha Ghodgaonkar, \\Shreya Ghosh, George K. Thiruvathukal\IEEEauthorrefmark{1}, Yung-Hsiang Lu}
    \IEEEauthorblockA{Purdue University, School of Electrical and Computer Engineering, West Lafayette, IN, USA}
    \IEEEauthorblockA{\IEEEauthorrefmark{1}Loyola University Chicago, Department of Computer Science, Chicago, IL, USA}
}
\title{Low-Power Object Counting with\\ Hierarchical Neural Networks}

\maketitle

\begin{abstract}

Deep Neural Networks (DNNs) can achieve state-of-the-art accuracy in many computer vision tasks, such as object counting. 
Object counting takes two inputs: an image and an object query
and reports the number of occurrences of the queried object.
To achieve high accuracy on such tasks, DNNs require billions of operations, making them difficult to deploy on resource-constrained, low-power devices. Prior work shows that a significant number of DNN operations are redundant and can be eliminated without affecting the accuracy. To reduce these redundancies, we propose a hierarchical DNN architecture for object counting. This architecture uses a Region Proposal Network (RPN) to propose regions-of-interest (RoIs) that may contain the queried objects. A hierarchical classifier then efficiently finds the RoIs that actually contain the queried objects. The hierarchy contains groups of visually similar object categories. Small DNNs are used at each node of the hierarchy to classify between these groups. The RoIs are incrementally processed by the hierarchical classifier. If the object in an RoI is in the same group as the queried object, then the next DNN in the hierarchy processes the RoI further; otherwise, the RoI is discarded. By using a few small DNNs to process each image, this method reduces the memory requirement, inference time, energy consumption, and number of operations with negligible accuracy loss when compared with the existing object counters.

\end{abstract}


\input{introduction.tex}

\input{related_work.tex}
\input{approach.tex}

\input{experimental.tex}
\input{conclusions.tex}

\vspace{-0.1cm}
\def\bibfont{\small}
\printbibliography
\end{document}

%% file: introduction.tex
\vspace{-0.2cm}
\section{Introduction}
\vspace{-0.1cm}
Deep Neural Networks (DNNs) are widely used in computer vision. Many computer vision tasks, such as image classification and object detection, use DNNs to achieve high accuracy~\cite{Yolo9000, FRCNN}. In image classification, the input is an image with a single object and the task is to identify the object. In object detection, the input is an image with multiple objects and the task is to localize and identify each object. Object counting is another computer vision task. This task takes two inputs:
(1)~image with multiple objects and (2)~an object category query.
The task reports the number of occurrences of the queried category of objects in the image~\cite{glance, where}. Object counting can be used in automatic traffic control~\cite{covid}, crowd management~\cite{icip}, and wildlife monitoring~\cite{where}. Most existing techniques require large DNNs that are compute and memory intensive to achieve high accuracy~\cite{lpirc}. The existing techniques are difficult to deploy on embedded devices where resources are scarce and energy efficiency is critical. Overcoming this difficulty will enable computer vision applications on mobile systems, drones, and wearable devices~\cite{8050296, vps}. 


\begin{figure}[t!]
        \centering
            \centering
            \includegraphics[width=0.35\textwidth]{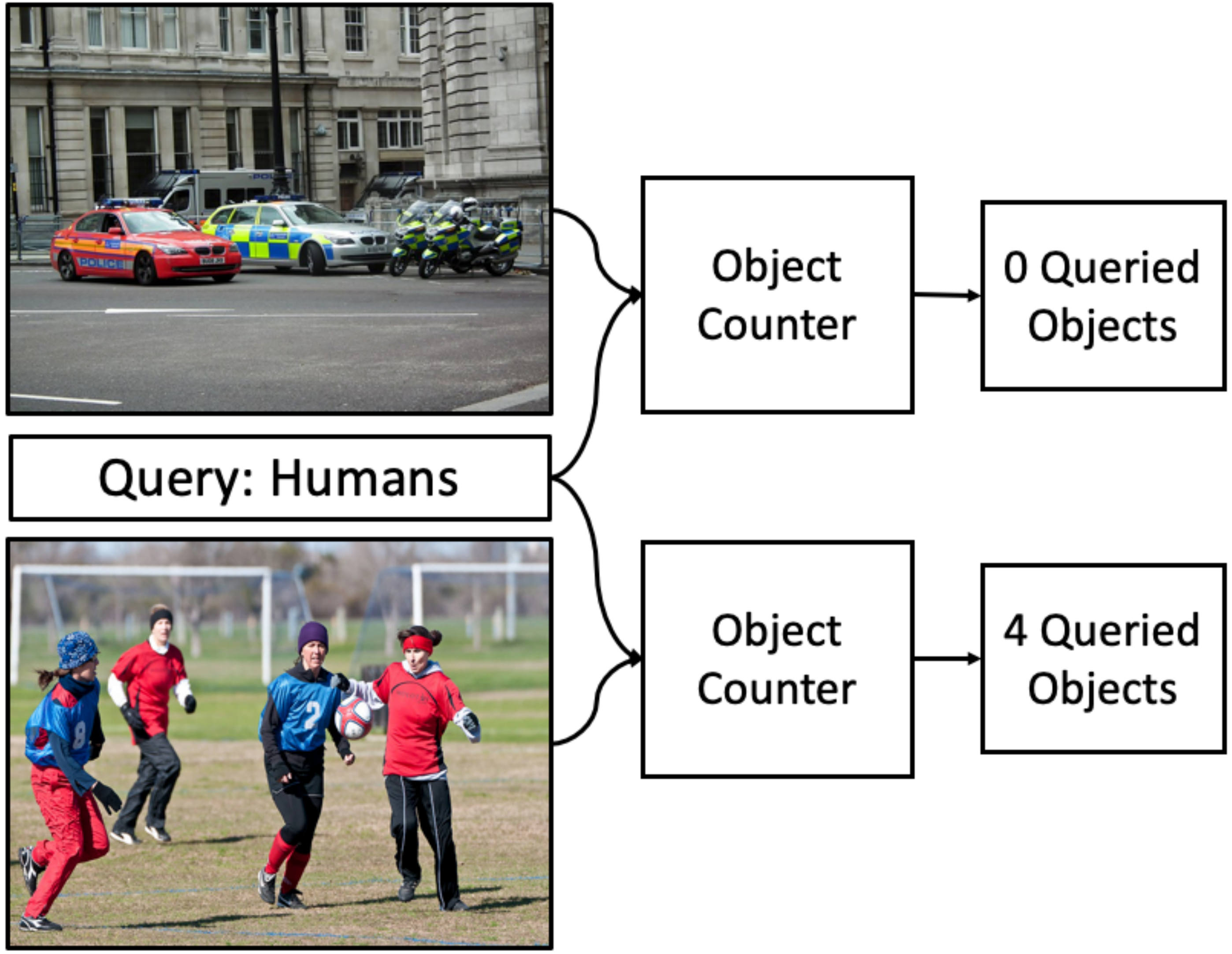}
        \caption{Two examples of object counting: when the query is ``Humans'', the image on top reports zero queried objects, and the image at the bottom reports four queried objects.}
        \vspace{-0.65cm}
        \label{fig:over}
\end{figure}

\begin{figure}[]
   \centering
   \subfloat[][]{\includegraphics[width=.20\textwidth]{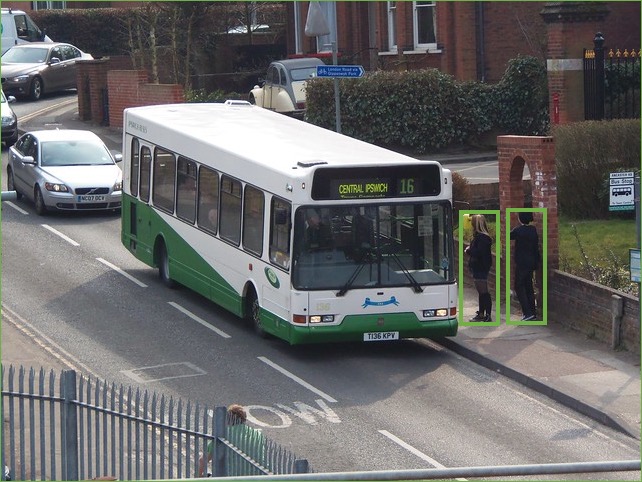}}\,
   \subfloat[][]{\includegraphics[width=.20\textwidth]{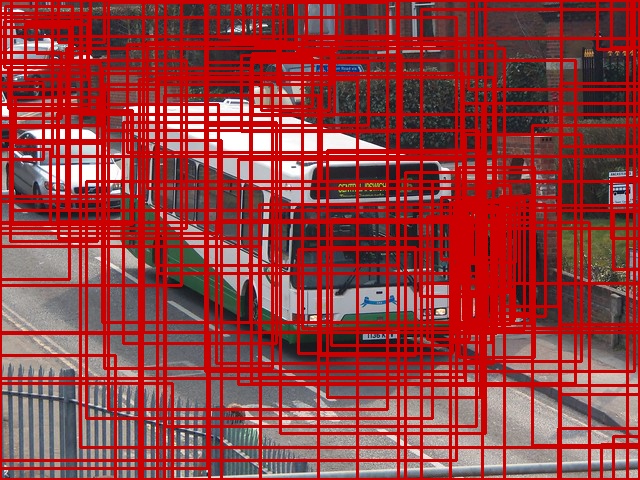}}
   \caption{(a) Image with two humans (marked in green). (b)~The RoIs proposed by the Region Proposal Network (marked in red). Existing techniques process all RoIs with large DNNs to detect all objects. The queried objects are counted.}
   \label{fig:roi_problem}
   \vspace{-0.65cm}
\end{figure}

Fig.~\ref{fig:over} describes the object counting problem with two examples. The inputs are an image and an object query; the output is the number of occurrences of the queried object. 
Existing object counters are based on object detectors~\cite{Yolo9000, FRCNN}. These techniques first propose regions-of-interest (RoIs) in an image. RoIs are areas of an image that potentially contain objects. These methods process all the RoIs with large DNNs to find all objects; then, the occurrences of the queried object are counted~\cite{glance}. As seen in Fig.~\ref{fig:roi_problem}, if the query object  is \textit{human}, existing methods first propose many RoIs (marked in red), then identify the objects in each RoI, and finally count the number of humans. In this example, there are about 6000 RoIs proposed and 2 humans. For counting the 2 humans, the computation involved in processing all the RoIs with large DNNs, to detect every object in the image is redundant. The redundancies should be avoided to improve energy-efficiency.

\begin{figure}[t!]
	\centering
		\includegraphics[width=0.4\textwidth]{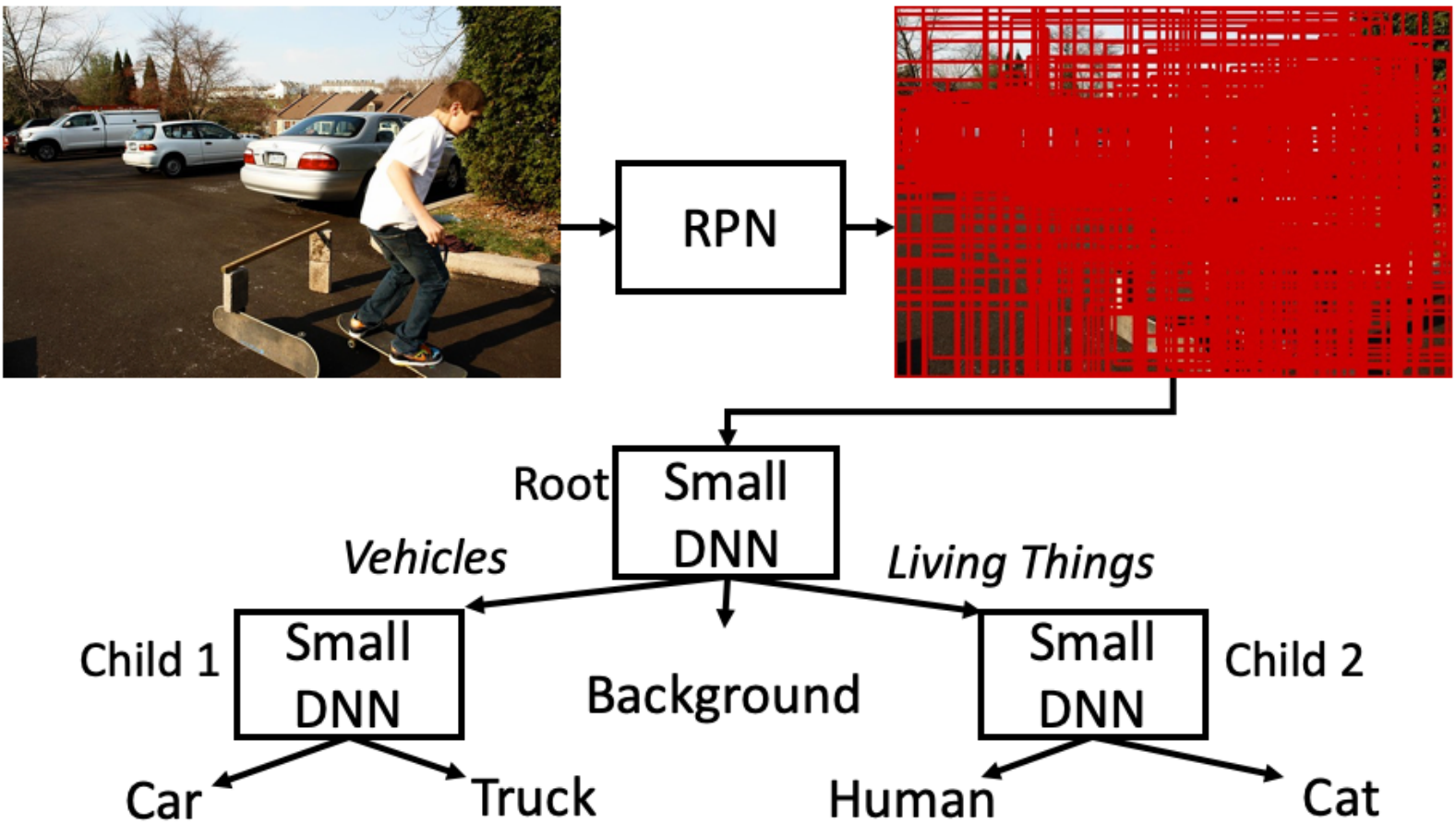}
		\caption{Proposed Technique: A RPN finds the RoIs in the image. The RoIs are then processed by the small DNN at the root. When the object query is \textit{car}, only the RoIs that get classified as vehicles by the root will be processed by the small DNN at Child~1 to distinguish between cars and trucks.}\label{fig:hier}
		\vspace{-0.6cm}
\end{figure}

This paper proposes a novel hierarchical DNN technique that reduces the redundant computation to perform low-power object counting. The object categories are grouped based on their visual similarity to form a hierarchy. Each node of the hierarchy contains a small DNN that distinguishes between its subsequent branches. This approach can be understood through the example shown in Fig.~\ref{fig:hier}. A RPN is used to find ROIs (shown with red bounding boxes) in the image. All the RoIs are processed by a small DNN at the root. When the object query is \textit{car}, only the RoIs containing vehicles are processed by child 1 for further classification. The other RoIs are not processed further. The DNN associated with child~2 is not used. By doing so, redundant operations are avoided.


The proposed method is evaluated against existing object counters. Our experimental results show that the proposed method saves 58\%-98\% memory, 19\%-95\% energy, 19\%-95\% inference time, and 45\%-90\% operations on an NVIDIA Jetson Nano using the Pascal VOC, and COCO datasets. These performance gains come with a 0.6 unit increase in Root Mean Square Error (counting error).




%% file: related_work.tex
\vspace{-0.1cm}
\section{Related Work}
\vspace{-0.1cm}
\textbf{Object Counting by Detection:} 
These methods first detect all the objects in an image and then count the number of instances of the queried object. Most object detectors are not suitable for low-power devices because they require large, power-hungry DNNs to determine the location and size of every object in the image. YOLO~\cite{Yolo9000}, Single Shot Detector (SSD)~\cite{SSD}, and SqueezeDet~\cite{squeezedet_2019} are single-stage object detectors. They are faster but less accurate than two-stage detectors like Faster RCNN (RPN + Classifier)~\cite{FRCNN}. Our method is a two-stage object counter that uses a RPN with a hierarchical DNN to reduce the redundant computation. We do not use single-stage detectors because they use very large DNNs, from which redundancies can not be removed easily. 


\textbf{Other Object Counting Techniques:}
Tu et al. use clustering to count objects based on hand-selected features~\cite{21}. Clustering methods often have lower accuracy compared with DNN-based approaches. Lempitsky et al.~\cite{8} use density estimation with a least-squares objective function. This technique can not count objects of different sizes because the density estimation kernel size is selected manually. Glance~\cite{glance} trains a large DNN with image-level labels with the counts of each object category; it requires a large power-hungry DNN to achieve high accuracy. Our method performs low-power object counting with hierarchical DNNs. 

\textbf{Low-Power Computer Vision:} Quantization methods~\cite{Han2015, Goel2018} reduce the memory requirement and energy consumption of DNNs, but have low accuracy. MobileNet~\cite{Mob} uses bottleneck layers to reduce the number of parameters and operations. These techniques are used for image classification~\cite{goel2020survey}. In contrast, our method uses a hierarchy of small DNNs to perform efficient object counting.

\begin{figure}[]
   \centering
   \subfloat[][]{\includegraphics[width=.12\textwidth]{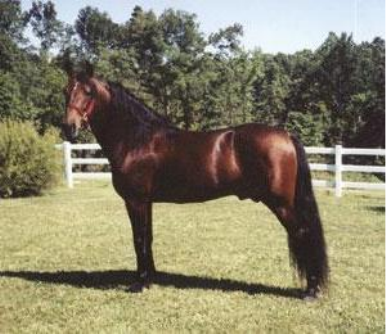}}\,
   \subfloat[][]{\includegraphics[width=.12\textwidth]{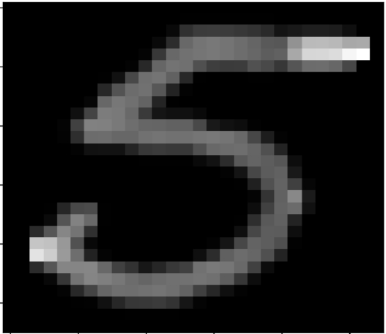}}\,
   \subfloat[][]{\includegraphics[width=.12\textwidth]{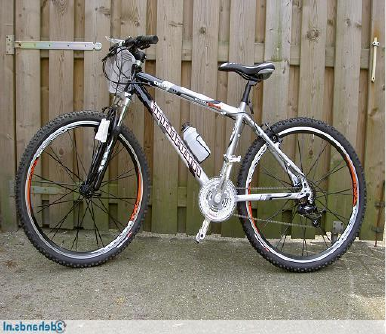}}\,
   \subfloat[][]{\includegraphics[width=.12\textwidth]{Figures/Picture6.pdf}}
   \caption{Most existing low-power image classifiers~\cite{Han2015, Goel2018} and hierarchical DNNs~\cite{G2008, FALCON, treecnn} are only compatible with images (a)-(c): a single object in the center of the frame. The hierarchical object counter presented in this paper can work for (a)-(c) and (d) with multiple objects of different categories.}
   \label{fig:sub1}
   \vspace{-0.6cm}
\end{figure}


%

\textbf{Hierarchical Computer Vision:}
Some techniques build hierarchical architectures for computer vision. There are three major techniques to build hierarchies: (1)~\textit{hierarchical clustering}: finds the distances between the centroids of categories, and then groups the two closest categories together~\cite{G2008}, (2)~\textit{semantic similarity}: uses conceptual and lexical relations between the different categories to group them together~\cite{Yolo9000, WordNet}, (3)~\textit{visual similarity}: uses DNN feature maps and layer activations to find similar categories~\cite{FALCON, treecnn}. 
These techniques can be used only for images with single objects in the center, such as Fig.~\ref{fig:sub1}(a)-(c) and 
cannot be used for object counting. Our method uses a tree-based architecture for object counting. It can work with images that contain multiple objects of different categories, as seen in Fig.~\ref{fig:sub1} (d).

{\bf Our Prior Work:} Our prior work~\cite{todaes} proposes a technique to use hierarchical DNNs for low-power image classification. 
To achieve high accuracy with low memory, computation, and energy requirements, the method uses the output of a DNN's softmax layer to identify and group visually similar categories. For example, birds and airplanes in the sky are visually similar, apples and bananas are visually dissimilar.
We also show that a fuzzy system based membership function is an effective method to group similar categories without the need for a manually selected threshold. The technique obtains higher accuracy than existing hierarchical image classifiers. Our prior method can only be used with image classification datasets, e.g. Fig~\ref{fig:sub1}(a)-(c). This paper improves our previous method by extending to images with multiple objects.

{\bf The contributions of this work include:} (1)~To the best of our knowledge, we are the first method to employ hierarchical DNNs for low-power object counting and can handle images that contain multiple objects.
(2)~Our method can group visually similar object categories into hierarchies, even when datasets contain images with multiple objects.
(3) Our method can systematically and efficiently construct the hierarchical object counter for large datasets and uses tree structures to reduce redundant computation.
(4) Our experiments show that the method consistently outperforms existing object counters in terms of memory, computation, and energy requirements.

%% file: approach.tex
\vspace{-0.1cm}
\section{Hierarchical Object Counting}
\vspace{-0.1cm}

This section explains the working of the proposed hierarchical object counter. Section~\ref{sub:cons} describes how to construct the hierarchical DNNs 
for  the  object counter. We then discuss the training process and how object counting is performed with the proposed hierarchical DNNs.

\vspace{-0.1cm}
\subsection{Constructing the Hierarchical Object Counter}
\label{sub:cons}
\subsubsection{Selecting DNN Architectures}
\label{subsub:nas}


The proposed method uses a Region Proposal Network (RPN) and a hierarchical classification DNN to efficiently count the queried objects. The sizes and architectures (number of layers and filters) of the DNNs need to be selected to achieve an acceptable tradeoff between accuracy and efficiency. The efficiency of DNNs is usually measured with the~\textit{information density} metric~\cite{acc_den}. The information density of a DNN is the ratio of its accuracy to its memory requirement.
The proposed method uses object detection to count objects and the accuracy of each DNN in the hierarchy is measured using Mean Average Precision (mAP). A higher mAP indicates that the predicted bounding boxes closely resemble the ground-truths.
The change in information density~\cite{acc_den} is used to compare the efficiency of two DNNs $D_i$ and $D_{i+1}$ with $i$ and $i+1$ layers, respectively. It is represented as $\Delta ID(D_i, D_{i+1}) = \frac{a_{i+1} - a_i}{m_{i+1} - m_i}$, where $a_i$ and $m_i$ are the mAP and memory requirement of $D_i$, respectively. 
Using $\Delta ID(D_i, D_{i+1})$ we can quantify whether the improvement of mAP by selecting $D_{i+1}$ instead of $D_i$ is worth the increase in memory requirement.


\begin{figure}[b!]
    \vspace{-0.3cm}
	\centering
		\includegraphics[width=0.49\textwidth]{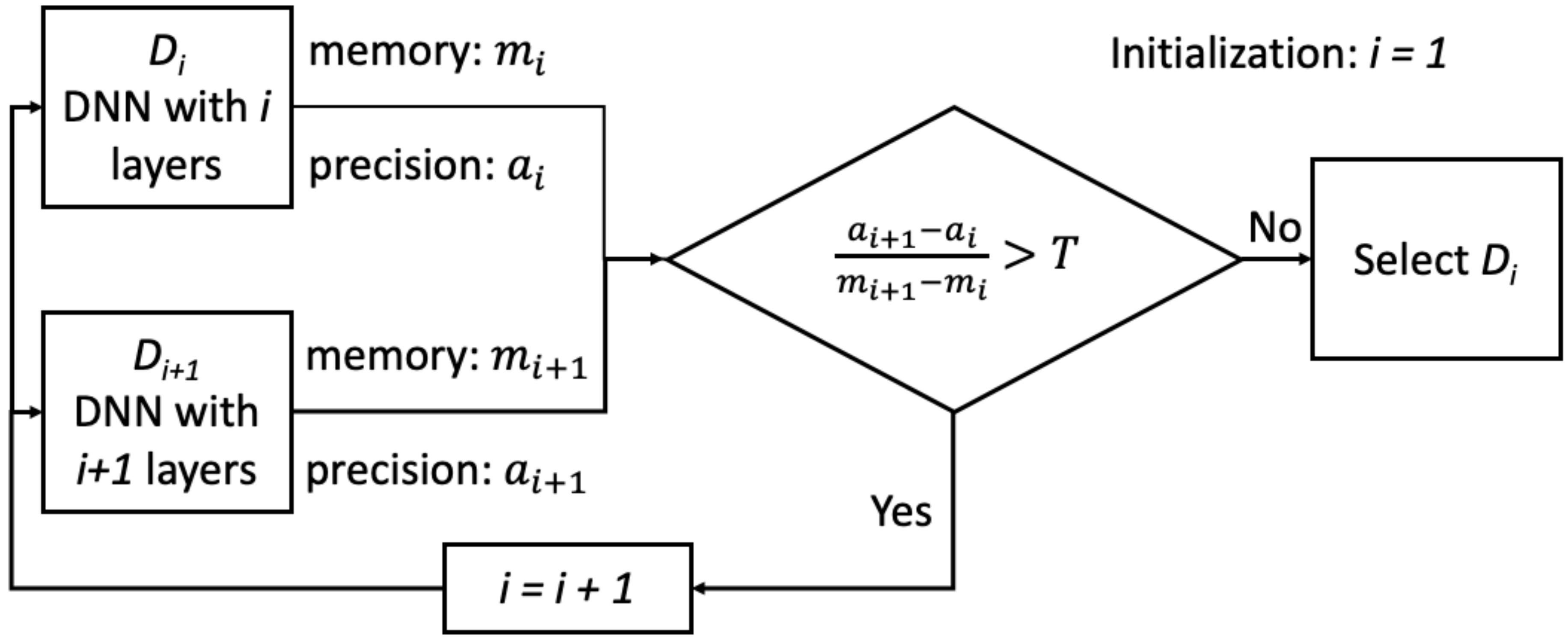}
		\caption{Proposed method to select DNN architectures in the hierarchical object counter.}\label{fig:PC}
		\vspace{-0.1cm}
		
\end{figure}


\begin{figure*}[h!]
	\centering
		\includegraphics[width=0.85\textwidth]{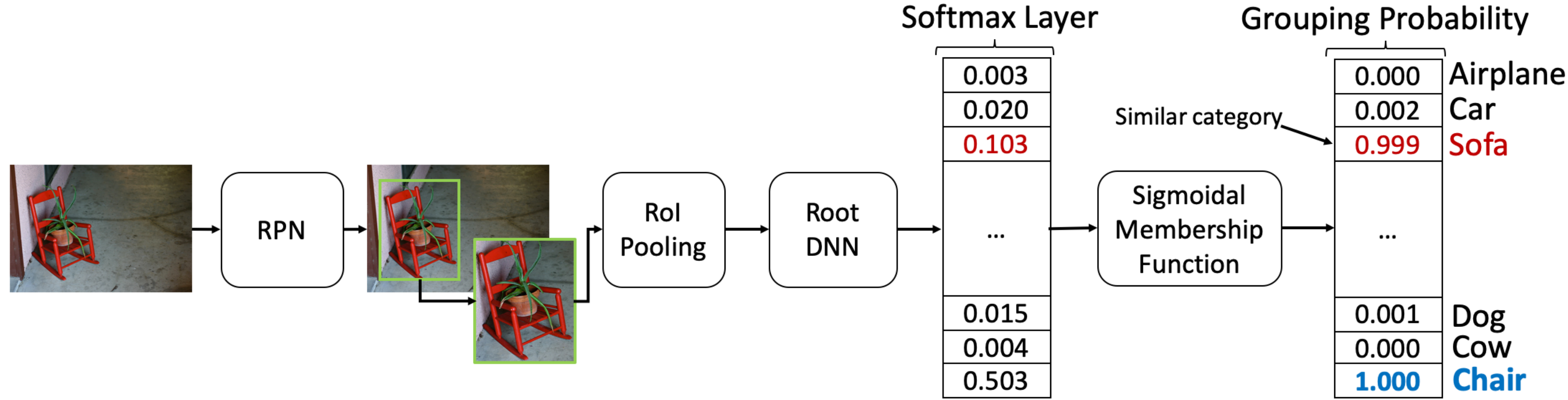}
		\caption{The Object Softmax Similarity method applied to the root node for the Pascal VOC dataset. The RPN proposes RoIs to isolate the objects in the image. The RoI-Pooling block resizes the RoIs such that they can be processed by the DNN. The softmax output of the DNN  is passed through the sigmoidal membership function to obtain the grouping probability.}\label{fig:OSS}
		\vspace{-0.52cm}
\end{figure*}

The algorithm for finding the architecture for each DNN in the hierarchy is given in Fig.~\ref{fig:PC}. This algorithm begins with $i = 1$. Two DNNs, with one ($i = 1$) convolutional layer and two ($i + 1$) convolutional layers, are trained to obtain their mAPs: $a_1$ and $a_2$, and memory requirements: $m_1$ and $m_2$ (in MB). 
The value of $\Delta ID(D_1, D_2)$ is then computed and compared against an empirical threshold, $T$. If the value of $\Delta ID(D_1, D_2) > T$, then this process continues for increasing values of $i$ until $\Delta ID(D_i, D_{i+1}) \le T$. This paper selects the value of $T = 0.001$ by experiments~\cite{todaes}. 
A small $T$ results in large DNNs because multiple layers are added. Since large DNNs can distinguish between categories effectively, the similar object grouping method (discussed in Section~\ref{subsub:sim}) results in a short tree. Such hierarchies 
consume more energy because they do not eliminate many redundant operations. A large value of $T$ leads to multiple small DNNs in a tall tree. Tall trees consume more energy because multiple small DNNs need to be fetched from memory.

\begin{table}[htb!]
    \centering
    \begin{tabular}{|c|c|c|}
        \hline
        \begin{tabular}[c]{@{}c@{}}Region Proposal\\ Network (RPN)\end{tabular} & \begin{tabular}[c]{@{}c@{}}Root DNN\end{tabular} & \begin{tabular}[c]{@{}c@{}}Child 1 DNN\end{tabular} \\ \hline
        Input: 300 $\times$ 425 $\times$ 3 & Input: 14 $\times$ 14 $\times$ 128  & Input: 7 $\times$ 7 $\times$ 128 \\ \hline
        \begin{tabular}[c]{@{}c@{}}conv3-32\\ conv3-32\\ conv3-32\end{tabular} & \begin{tabular}[c]{@{}c@{}}conv3-128\\ conv3-128\\ conv3-128\end{tabular} & \begin{tabular}[c]{@{}c@{}}conv3-128\\ conv3-128\\ conv3-128\end{tabular}  \\
        Max Pool 2 $\times$ 2 & Max Pool 2 $\times$ 2 & Max Pool 2 $\times$ 2 \\
        \begin{tabular}[c]{@{}c@{}}conv3-64\\ conv3-64\\ conv3-64\end{tabular} & Linear 6272 $\times$ 13 & conv3-128 \\
        Max Pool 2 $\times$ 2 &  &  Linear 1152 $\times$ 17 \\
        \begin{tabular}[c]{@{}c@{}}conv3-128\\ conv3-128\\ conv3-128\end{tabular} &  & \\
        RoI-Pool 7 $\times$ 7 & & \\
        \hline
        
    \end{tabular}
    \caption{Examples of DNN architectures obtained for the Pascal VOC dataset with our method. The Region Proposal Network has 9 convolutional layers. The root DNN contains 3 convolutional layers and one fully connected output layer. The first child DNN contains 4 convolutional layers. The other children DNNs have similar architectures to the Child 1 DNN.}
    \label{tab:arch}
    \vspace{-0.3cm}
\end{table}

To build the hierarchical object counter, the architecture of the RPN is chosen first, then the architectures of the DNNs in the hierarchy are selected. 
The time taken to find DNN architectures is significant because multiple DNNs of different sizes have to be trained and evaluated. To reduce the training time, we use an image sampling technique~\cite{zoph2018} to find a subset of the training dataset. We also employ a learning curve extrapolation algorithm~\cite{extra} to estimate the mAP of fully-trained DNN after just 5 training epochs. These approximation strategies are suitable for evaluating tradeoffs when selecting the DNN size because the $\Delta ID(D_i, D_{i+1})$ uses the difference in mAP between DNNs when the training process is the same~\cite{extra}. TABLE~\ref{tab:arch} shows the architectures of the DNNs obtained with the proposed method. The RPN contains 9 convolutional layers, the root DNN contains 3 convolutional layers and one fully connected output layer. Max Pool layers are used after every three convolutional layers to reduce the feature size. The output feature map of the root DNN is the input to the child DNN.


\subsubsection{Grouping Visually Similar  Categories}
\label{subsub:sim}

In order to construct the hierarchical DNNs, visually similar categories need to be grouped together in the form of a tree. The existing techniques to build hierarchies based on visual similarities only work for image classification (each image has only one object); they cannot be used for object counting (each image has multiple objects)~\cite{FALCON, treecnn}. 
To solve this problem, this paper uses
a new method to find the visual similarities between object categories: Object Softmax Similarity (OSS).


OSS is applied to every node of the hierarchy.
It is first applied to the root of the hierarchy to find the first level of children, as described in Fig.~\ref{fig:OSS}. 
The images from the training dataset are fed into the RPN (architecture is selected using the technique described in Section~\ref{subsub:nas}) to isolate the individual objects.
Each RoI output by the RPN is then labeled with the category of the object that it contains. We follow the common practice and label the RoIs that have a 70\% overlap with a ground-truth bounding box~\cite{FRCNN}. The other RoIs are considered to be part of the background. OSS uses the softmax output of a DNN to evaluate if two categories are visually similar. To  process RoIs with a DNN, they must all have the same dimensions (height and width). To resize the RoIs into a fixed size without distorting its features, a technique called RoI-Pooling is performed~\cite{FRCNN}. The resized RoIs are then processed by the DNN at the root of the hierarchy after the architecture is selected using the technique described in Section~\ref{subsub:nas}. The DNN's softmax output is used to quantify the similarity between categories. For each RoI containing an object belonging to category $A$, the DNN's softmax outputs are accumulated: $\sum softmax(A)$. After processing every image in the dataset, the accumulated softmax values are divided by $|A|$, the number of RoIs labeled as $A$ in the entire dataset, to find the averaged softmax output for category $A$. Since this averaged softmax output contains a value corresponding to every object category, it is possible to identify the categories similar to category $A$. If the value corresponding to category $B$ is large, the DNN is confused about the categories $A$ and $B$. This confusion is caused by visually similar categories. Visually similar categories are grouped together into a new node in the hierarchy using the sigmoidal membership function. For every newly formed node, the same process is repeated: the DNN architecture is selected and the similarity of between its children is computed using OSS. 

OSS uses a fuzzy system-based approach to determine if the averaged softmax value is large enough for two object categories to be grouped. 
A sigmoidal membership function is used to assign a probability for each object category being grouped with every other object category~\cite{zadeh}. The use of this function ensures that visually similar categories $A$ and $B$ will be grouped with a high probability ($\approx$ 1), and visually dissimilar categories will not be grouped. For example, as seen in Fig.~\ref{fig:OSS}, \textit{sofa} has a high softmax output and will be grouped with \textit{chair}, while \textit{airplane} will not be grouped with \textit{chair}. The hierarchy shown in Fig.~\ref{fig:hier1} is obtained with the OSS method. The tree has a depth~=~4 and an average branching factor~=~3 for the Pascal VOC dataset, and depth~=~4 and an average branching factor~=~6 for the COCO dataset.





\vspace{-0.1cm}
\subsection{Training Method}

Each DNN in the hierarchical object counter needs to be trained with labeled data from the training dataset. Each image is annotated with bounding boxes and labels. 
The RPN is first trained using back-propagation. 
The DNNs in the hierarchical classifier are trained in a root-down fashion (root DNN is trained first). 
The DNNs at a given depth of the hierarchy are completely independent of one another. They are trained in parallel to reduce the total training time. Two loss functions are employed during training: Smooth L1 Loss for predicting bounding boxes and Categorical Cross-Entropy Loss for identifying the objects in each RoI.

\vspace{-0.1cm}
\subsection{Performing Object Counting}

\begin{figure*}[t!]
	\centering
		\includegraphics[width=0.75\textwidth]{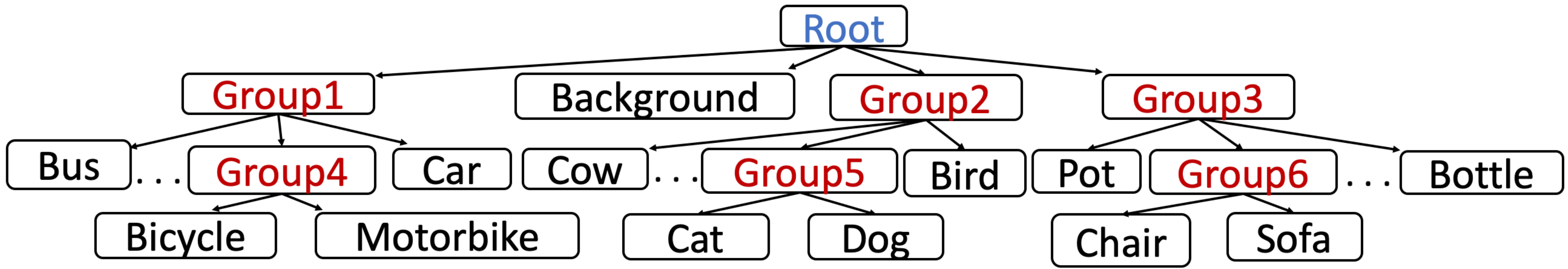}
		\caption{Part of the hierarchy obtained with the Object Softmax Similarity (OSS) method for the Pascal VOC dataset. The number of children under each node is dynamically determined during the training process. }\label{fig:hier1}
		\vspace{-0.4cm}
\end{figure*}

When performing object counting, the inputs are an image and a query. First, the RPN and the DNNs of the hierarchy on the path from the root to the queried object category are loaded into memory.
The input image is processed by the Region Proposal Network (RPN) to find RoIs. The RoIs are processed by the small DNN at the root of the hierarchical classifier. The root DNN classifies each RoI into groups of similar categories. As seen in Fig.~\ref{fig:hier1}, the DNN at the root classifies each RoI as Group 1, Group 2, Group 3, or background. The RoIs that are classified into the group that contains the queried object category are processed further by the corresponding child DNN. The other RoIs are discarded. This process continues: the RoIs get classified into smaller sub-groups within the previously identified groups. Only the RoIs in the same group as the queried object category are processed further. Non-Maximal Suppression~\cite{Yolo9000} is performed at the end to eliminate overlapping RoIs. The number of remaining RoIs is the count of the queried object category.

The output feature map (output of the last convolutional layer) of the parent DNN is used as the input to the chosen child DNN in the hierarchy. This ensures that the computation performed at the parent is not repeated by the child, thus reducing redundant computation. The children act like specialized extensions of the parent. This is the reason why the children DNNs are small even though they are classifying between visually similar object categories.




%% file: experimental.tex
\vspace{-0.1cm}
\section{Experiments and Results}
\vspace{-0.1cm}
\subsection{Datasets Used}
We use two image datasets: Pascal VOC and COCO. Pascal VOC contains 11,540 color images with 20 different categories. COCO contains over 80,000 training images with 80 categories. The objects in these datasets are annotated with ground-truth object detection bounding boxes.

\vspace{-0.1cm}
\subsection{Experimental Setup}

All the DNNs of the proposed hierarchical object counter are trained using gradient descent. A batch size of 4 is used for both datasets, for 20 epochs. An initial learning rate of 0.004 is used and is dropped by a factor of 10 every 8 epochs. The training time for the proposed architecture includes the time taken to find the visual similarities and DNN architectures. When using the NVIDIA TITAN X GPU, the total training time for the Pascal VOC dataset is approximately 48 hours, which is $1.5\times$ the training time of existing techniques.

The memory requirement and number of operations are found using the \texttt{torchsummary} and \texttt{thop} PyTorch libraries, respectively. The Yokogawa WT310E Power Meter is used to measure the energy consumption when the object counters are deployed on the NVIDIA Jetson Nano. 

\vspace{-0.1cm}
\subsection{Results}

The proposed method is compared with the state-of-the-art object counters in terms of memory requirement, number of operations, and energy requirement. The accuracy is measured with the root mean squared error (RMSE) and mean average precision (mAP) obtained on the testing dataset (not used for training).
We include YOLO v3~\cite{Yolo9000}, Faster RCNN~\cite{FRCNN}, SqueezeDet~\cite{squeezedet_2019}, MobileNet-SSD~\cite{SSD}, LC-FCN with ResNet~\cite{where}, and Glance~\cite{glance} in our experiments. We also include the ``Semantic Tree'': our hierarchical object counter with semantic similarities~\cite{WordNet}, instead of the Object Softmax Similarity (OSS) method. This is done to show the need for hierarchies based on visual similarities.

TABLE~\ref{Tab:comp} compares the model size, number of operations, RMSE, and mAP of the different techniques. For the proposed method, the model size is reported as the sum of the model sizes of the RPN and the DNNs along the longest path from the root to a leaf. The proposed hierarchical object counter requires the least memory. When compared with Faster RCNN on Pascal VOC, our method requires 98.54\% $(1 - \frac{16}{1100} = 0.9854)$  less memory. 
Smaller model sizes require fewer memory accesses, leading to faster inference and lower energy consumption.

The number of operations for our method is the number of operations in RPN and the DNNs along the longest path from the root to a leaf. Our method reduces redundancies by using a subset of the DNNs for every input; the number of operations is significantly reduced when compared with existing techniques. It requires 87.50\% $(1 - \frac{42}{336} = 0.8750)$  fewer operations than MobileNet-SSD on the COCO dataset.
Only Tiny YOLO requires fewer operations. Our method significantly outperforms Tiny YOLO in terms of accuracy and memory requirement.

\begin{table}[]
\centering
\begin{tabular}{|p{0.5cm}|p{2cm}|r|r|r|r|}
\hline
 \multicolumn{1}{|c|}{Dataset} & \multicolumn{1}{c|}{Technique} & \multicolumn{1}{c|}{\begin{tabular}[c]{@{}c@{}}Model\\Size\end{tabular}} & \multicolumn{1}{c|}{\begin{tabular}[c]{@{}c@{}}Number of\\ Operations\end{tabular}} & \multicolumn{1}{c|}{\begin{tabular}[c]{@{}c@{}}Test\\ RMSE\end{tabular}} & \multicolumn{1}{c|}{\begin{tabular}[c]{@{}c@{}}Test\\ mAP\end{tabular}} \\ \hline
\multirow{9}{*}{\begin{tabular}[c]{@{}c@{}}Pascal\\VOC\end{tabular}}
& Faster RCNN~\cite{FRCNN} & 1,100 & 440 B &  1.38  &  0.762 \\
 & SSD~\cite{SSD}& 39 & 336 B &  1.73  &  0.724\\
 & YOLO v3~\cite{Yolo9000} & 248 & 141 B &  1.61  & 0.734 \\
 & Tiny YOLO~\cite{Yolo9000} & 55 & 5 B &  2.32 &  0.300\\
 & SqueezeDet~\cite{squeezedet_2019} & 57 & 77 B &  2.01  & 0.643 \\
& Glance~\cite{glance} & - & - &  1.87  & - \\
 & LC-FCN~\cite{where} & 194 & 156 B &   1.20  & - \\
 & Semantic Tree & 16 & 41 B &  2.56 &  0.211\\
& \textbf{Our Method} & 16 & 42 B &  1.80  &  0.668\\ \hline
\multirow{8}{*}{COCO}
& Faster RCNN~\cite{FRCNN} & 1,100 & 440 B &  1.99 &  0.591\\
 &  SSD~\cite{SSD} & 40 & 336 B &  2.17 & 0.504 \\
 &YOLO v3~\cite{Yolo9000} & 249 & 141 B &  2.07  &  0.579 \\
 & Tiny YOLO~\cite{Yolo9000} & 55 & 5 B &  2.32 &  0.237 \\
 & SqueezeDet~\cite{squeezedet_2019} & 57 & 78 B &  2.99  &  0.507\\
 & Glance~\cite{glance} & - & - &   2.32  & - \\
 &  Semantic Tree & 20 & 44 B &  3.51  & 0.134  \\
 & \textbf{Our Method} & 19 & 44 B &  2.24  & 0.522 \\ \hline
\end{tabular}
\caption{The memory requirement (in MB), number of operations, Root Mean Squared Error (RMSE), and Mean Average Precision (mAP) for different techniques and datasets. ``-'' indicates that source code/data is not available}
\label{Tab:comp}
\vspace{-0.65cm}
\end{table}

\begin{figure}[]
	\centering
	\includegraphics[width=0.43\textwidth]{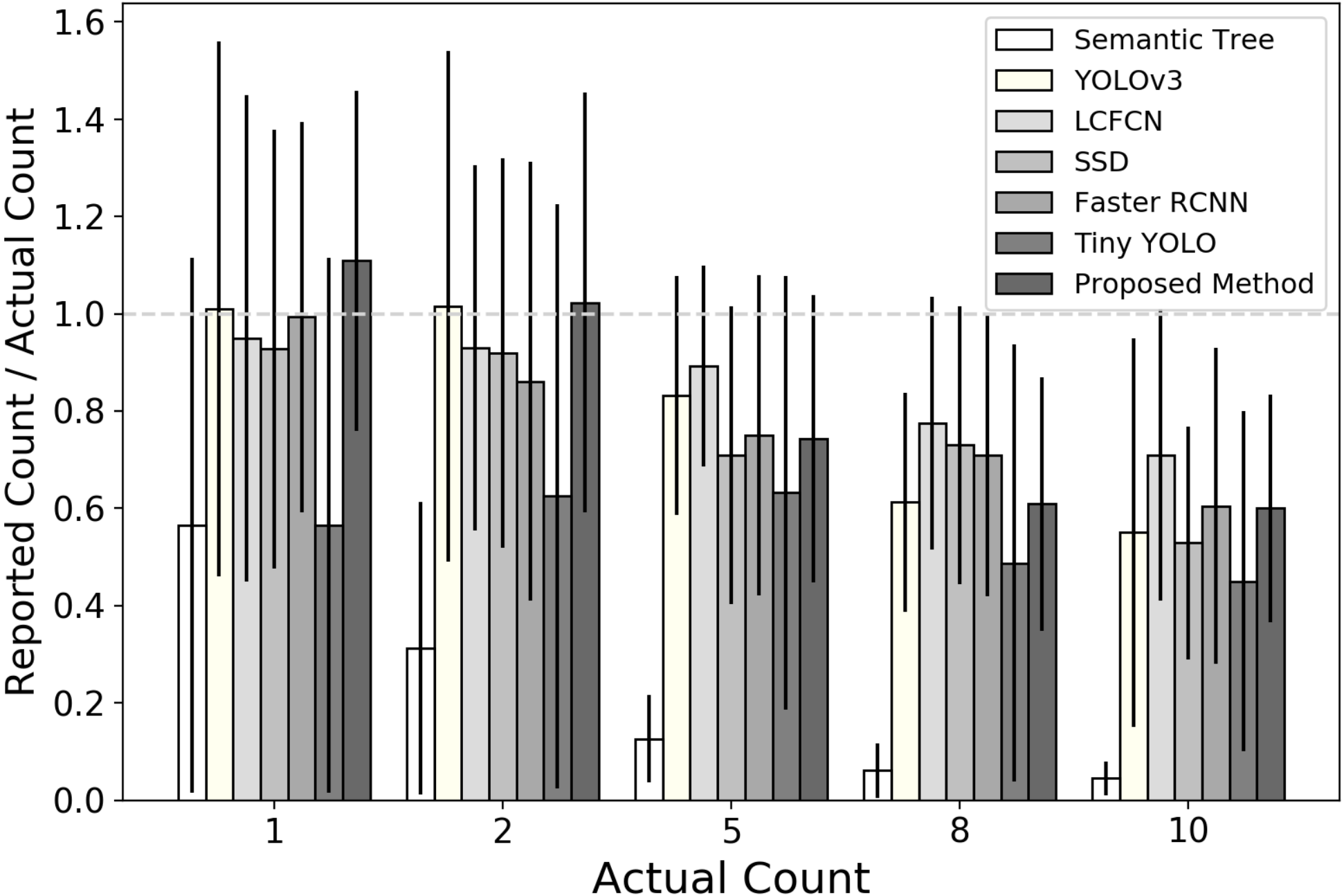}
	\caption{Normalized reported object count with increasing numbers of objects per image on Pascal VOC testing set.
	}
	\label{fig:count}
	\vspace{-0.22cm}
\end{figure}

\begin{figure}[t!]
	\centering
	\includegraphics[width=0.49\textwidth]{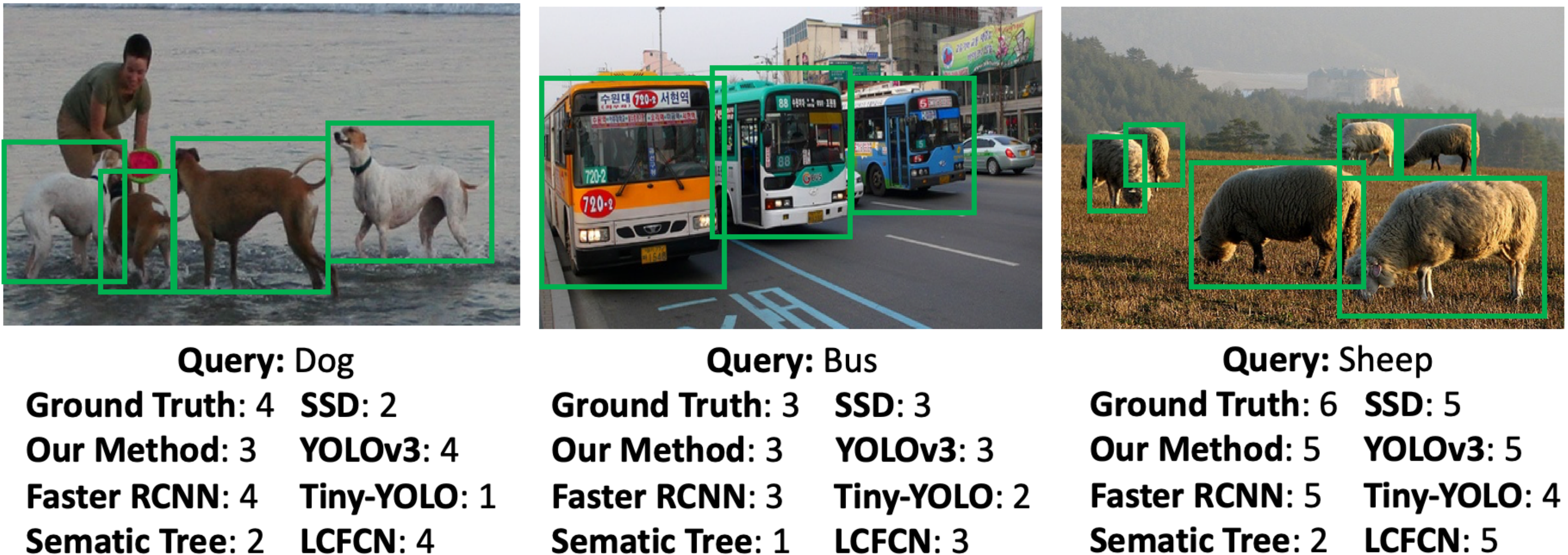}
	\caption{Output of different techniques on examples from the Pascal VOC dataset. Ground truth boxes are marked in green.}
	\label{fig:examples}
	\vspace{-0.12cm}
\end{figure}

The non-zero RMSE and mAP (reported in TABLE~\ref{Tab:comp}) are a commonly used metrics to measure the accuracy of object counting systems~\cite{glance}. The performance of the hierarchical object counter is comparable with the existing state-of-the-art techniques. The proposed technique outperforms SqueezeDet, Glance, and Tiny YOLO. Our hierarchical counter with OSS also outperforms the hierarchical counter with semantic similarities. This is because visually similar objects are sometimes semantically dissimilar (e.g. bird and airplane in the sky). The small DNNs in the hierarchy misclassify these objects, leading to poor counting accuracy. Fig.~\ref{fig:count} shows the normalized reported count (ratio of the reported count to the actual count) for images with different numbers of objects. When this ratio~=~1, the object counter obtains the best result.
Tiny YOLO and the Semantic Tree have the poorest accuracy. The proposed method obtains accuracy comparable with the state-of-the-art object counters. Fig.~\ref{fig:examples} shows object counting examples from the Pascal VOC dataset. The proposed technique reports counts close to the ground truth counts in all images.

TABLE~\ref{tab:pi} depicts the energy consumption and inference time of the PyTorch implementations of different object counters on an NVIDIA Jetson Nano, averaged over 500 images. The energy consumption for the proposed method is 95\% lower than YOLOv3, and 19\% lower than Tiny YOLO. The hierarchical object counter requires 19\%-95\% less inference time. This comparison is limied to YOLOv3 and Tiny YOLO because the other DNN architectures are too large to run on the NVIDIA Jetson Nano (they cause memory errors).


\begin{table}[t!]
\centering
\begin{tabular}{|c|c|r|r|r|}
\hline
Dataset & Metric & \multicolumn{1}{c|}{YOLOv3} & \multicolumn{1}{c|}{\begin{tabular}[c]{@{}c@{}}Tiny\\ YOLO\end{tabular}} & \multicolumn{1}{c|}{\begin{tabular}[c]{@{}c@{}}Our\\ Method\end{tabular}} \\ \hline
\multirow{2}{*}{\begin{tabular}[c]{@{}c@{}}Pascal\\ VOC\end{tabular}} & Inference Time & 22.33 & 1.25 & 1.02 \\ \cline{2-5} 
 & Energy Consumption & 162.00 & 9.90 & 8.10 \\ \hline
\multirow{2}{*}{COCO} & Inference Time & 24.12 & 1.45 & 1.12 \\ \cline{2-5} 
 & Energy Consumption & 175.00 & 11.46 & 8.92 \\ \hline
\end{tabular}
\caption{Energy consumption (J/image) and inference time (sec/image) comparison on an NVIDIA Jetson Nano.}
\label{tab:pi}
\vspace{-0.6cm}
\end{table}

%% file: conclusions.tex
\vspace{-0.1cm}
\section{Conclusions}


In this paper, we propose a novel low-power object counting method that uses hierarchical DNNs.
Our method uses several small DNNs (in the form of a tree) that work together for object counting. We demonstrate the efficacy of this approach using the NVIDIA Jetson Nano, an entry-level board with limited processing power and memory. We demonstrate that using small DNNs is energy-efficient, both in terms of operations required and the memory footprint, making it viable in low-power environments. Beyond energy-efficiency, inference is also faster with our hierarchical object counter.
We achieve these performance gains with negligible loss of accuracy. This is made possible via  a systematic method that identifies and groups visually similar categories into a hierarchy. 
